# Effective Cross-Utterance Language Modeling for Conversational Speech Recognition

Bi-Cheng Yan[1], Hsin-Wei Wang[1], Shih-Hsuan Chiu[1], Hsuan-Sheng Chiu[2], Berlin Chen[1]
[1] Department of Computer Science and Information Engineering, National Taiwan Normal University, Taipei, Taiwan
[2] Chunghwa Telecom Laboratories, Taoyuan, Taiwan
{bicheng, hsinweiwang, shchiu, berlin}@ntnu.edu.tw, samhschiu@cht.com.tw

*Abstract*—Conversational speech normally is embodied with loose syntactic structures at the utterance level but simultaneously exhibits topical coherence relations across consecutive utterances. Prior work has shown that capturing longer context information with a recurrent neural network or long short-term memory language model (LM) may suffer from the recent bias while excluding the long-range context. In order to capture the long-term semantic interactions among words and across utterances, we put forward disparate conversation history fusion methods for language modeling in automatic speech recognition (ASR) of conversational speech. Furthermore, a novel audio-fusion mechanism is introduced, which manages to fuse and utilize the acoustic embeddings of a current utterance and the semantic content of its corresponding conversation history in a cooperative way. To flesh out our ideas, we frame the ASR *N*-best hypothesis rescoring task as a prediction problem, leveraging BERT, an iconic pre-trained LM, as the ingredient vehicle to facilitate selection of the oracle hypothesis from a given *N*-best hypothesis list. Empirical experiments conducted on the AMI benchmark dataset seem to demonstrate the feasibility and efficacy of our methods in relation to some current top-of-line methods. The proposed methods not only achieve significant inference time reduction but also improve the ASR performance for conversational speech.

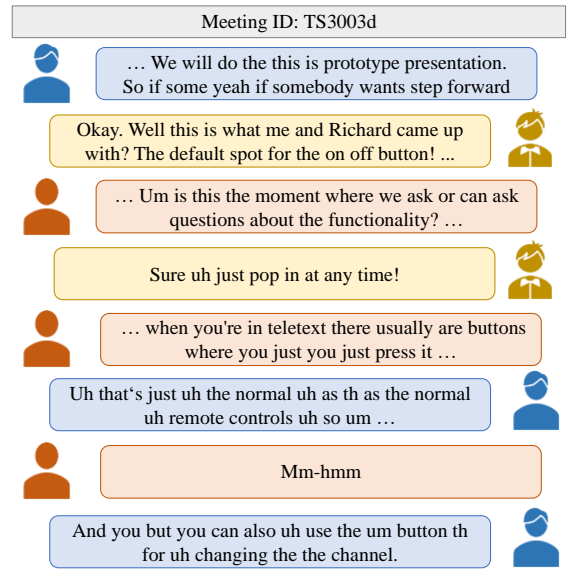

Fig.1. A snippet of a meeting conversation. This conversation was randomly chosen from the evaluation set of the AMI dataset, where three speakers participated in the meeting.

## I. INTRODUCTION

Recently, automatic speech recognition (ASR) has achieved remarkable success and reached human parity, thanks to the synergistic breakthroughs in neural model architectures and training algorithms [1][2]. Despite the great progress, current ASR systems are vulnerable to performance degradation for conversation-related applications such as digital personal assistants, smart speakers, interactive voice responses (IVR) and among others. The difficulties faced by conversational ASR is due mainly to its free-style speaking traits and long-term contextual dependencies across multiple utterances or speaker turns [3][4][5]. Fortunately, topical coherences and frequently repeated words often reside in consecutive utterances. A simple and lightweight approach to dealing with such kind of conversational ASR is to endow a language model (LM) with the ability to capture long-span lexical/semantic dependencies among words or utterances, and in turn leverage it to rerank the initial hypotheses generated by an ASR system. As an example, Figure 1 shows a conversation snippet culled from the AMI meeting corpus, which involves three participants and discusses about a remote control design project.

*N*-best hypothesis reranking is deemed to be a plug-and-play method for enhancing the performance of conversational ASR, which aims to find the oracle hypothesis that has the lowest word error rate from a given *N*-best hypothesis list. Most of the recent studies on cross-utterance language modeling have focused mostly on effective uses of recurrent neural network (RNN), long short-term memory (LSTM) and Transformer-based language models, to name just a few, to compute an updated LM score for each hypothesis in an autoregressive (AR) manner, calculating the probability of each hypothesis based on a left-to-right probabilistic chain rule for the reranking purpose. In [6], an LSTM-based neural cache mechanism was proposed, which encodes consecutive words occurring across historical utterances into a fixed-length vector representation. The vector representation is in turn used to adapt a given neural LM. In addition, a so-called dynamic evaluation method endeavored to fit an LSTM LM in the recent historical sentences via gradient descent-based adaptation [7]. Yet more recently, an LSTM-based fusion module was introduced to modulate transformer-based LMs [8][9] for better modeling power. However, since the aforementioned methods operate in an autoregressive manner, a practical drawback of them is that the inference time will grow linearly (for RNN and LSTM) or quadratically (for Transformer) with the input sequence [10][11], which would make these LM methods infeasible in many use cases. In addition, RNN and LSTM inherently suffer from the sharp nearby, fuzzy far away issues [12] and might fail to capture complex global structural dependencies among these utterances.

Building on these observations, we in this paper present a novel non-autoregressive LM approach that seeks to integrate cross-utterances, long-term word interaction cues into ASR *N*- best hypothesis reranking with two disparate fusion strategies: early fusion and late fusion. The former performs input augmentation, which simultaneously considers within-sentence and cross-sentence interactions among words for hypothesis reranking. The latter, in contrast, renders the semantic relevance between the conversation history and the holistic embedding of each hypothesis. Furthermore, a novel audio-fusion mechanism is introduced, developed to fuse and utilize the acoustic embeddings of a current utterance in conjunction with the semantic content of its corresponding conversation history. To turn the idea into reality, the ASR *N*-best hypothesis rescoring task is framed as a prediction problem, leveraging BERT (short for bidirectional encoder representations from Transformers [13]), an iconic pre-trained LM, as the fundamental vehicle to facilitate selection of the oracle hypothesis from a given *N*-best hypothesis list. The main contributions of this paper are summarized as follows:

- We put forward a non-autoregressive (NAR) LM approach for ASR *N*-best hypothesis reranking, which behaves as a lightweight and effective alternative to the current mainstream ones that operate in an autoregressive manner. The component models of our approach are trained on an objective that is tailored to the *N*-best hypothesis reranking task.
- We simultaneously take into account cross-utterance conversation context and audio (acoustic) features of a current utterance for ASR *N*-best hypothesis reranking, whereas most of the LM methods for this purpose are merely trained and evaluated on context that is restricted to the textual information of single utterances or their *N*-best hypotheses which are inadequate for dealing conversational ASR.
- Extensive experiments are conducted on the AMI meeting transcription benchmark corpus to validate the effectiveness and feasibility of our methods in comparison to several top-of-the-art reranking methods.

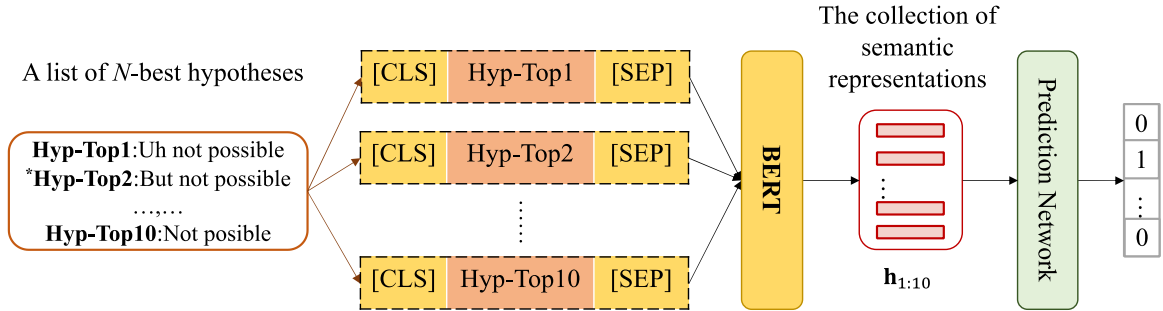

Fig.2. A schematic depiction of prediction BERT-based (PBERT) reranking model.

The rest of the paper is organized as follows. Section II sheds light on the proposed methodology for *N*-best list reranking. The experimental setup and results are reported in Section III, followed by the conclusion and future research directions in Section IV.

## II. PROPOSED METHODOLOGY

### A. BERT-based ASR N-best Hypothesis Reranking

Recently, recurrent neural LMs, such as those stemming from RNN, LSTM [14][15] and others [16], have enjoyed much success on a wide variety of ASR tasks. These LMs in general are hardly be used at the first-pass decoding stage of ASR due to the exponential growth of hypothesis search space with the increase in the modeling context of an LM. As such, an alternative and lightweight therapy is to employ them to compute an LM score for each hypothesis at the second-pass (viz. *N*-best hypothesis reranking) stage, making most ASR modules remain unchanged. However, recurrent neural LMs and the like usually operate in an autoregressive manner, which would be computationally expensive and hinders them from being used in production systems.

Departing from these methods, an effective BERT-based modeling framework for ASR *N*-best hypothesis reranking has been prototyped in our recent preliminary studies [17][18][19]. This framework aims to predict a hypothesis that would have the lowest WER (i.e., the oracle hypothesis) from an *N*-best list (denoted by PBERT) in a one-shot, non-autoregressive (NAR) manner. In realization, PBERT consists of two model components, namely BERT stacked with an additional prediction layer which is a simple fully-connected feedforward network (FFN) for oracle hypothesis prediction, as graphically illustrated in Figure 2. For a given ASR *N*-best hypothesis list, each hypothesis of the list is respectively taken as an input to the BERT component, and meanwhile [CLS] and [SEP] tokens are inserted at the beginning and end of each hypothesis, respectively. In turn, the resulting embedding vector of [CLS] is used as a semantic-aggregated representation of the input hypothesis. The [CLS] embedding vectors of all the *N*-best hypotheses are further augmented with their respective initial ASR decoding scores and consequently spliced together to be fed into the FFN component to output a prediction score (with the Softmax normalization) for each hypothesis.

Given a set of training utterances, each of which is equipped with an *N*-best hypothesis list generated by ASR and the indication of the oracle hypothesis that has the lowest WER, we can train the FFN component and fine-tune the BERT component accordingly. We should also note here that there have been some very recent studies that applied BERT as a mask-predict LM to rescore an ASR hypothesis in terms of the so-called pseudo-log-likelihood (PLL) score for *N*-best hypothesis reranking (denoted by MBERT for short)

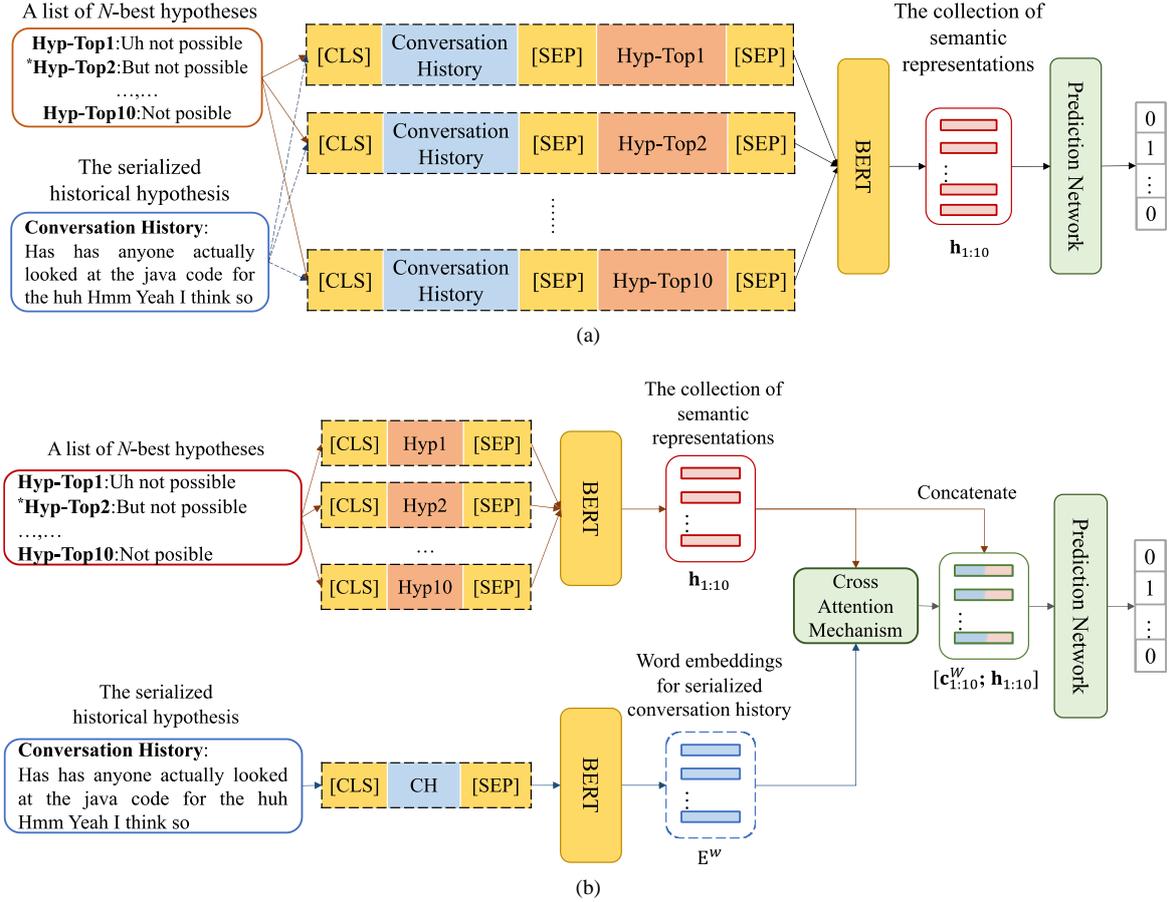

Fig.3. A schematic depiction of disparate BERT-based reranking strategies: (a) the EHPBERT model; (B) the LHPBERT model

[20][21][22]. However, unlike PBERT, MBERT processes each hypothesis in an isolated and autoregressive manner, and its speed is normally orders of magnitude lower than PBERT.

### B. Fusion of Cross-Utterance History Information

For conversational ASR tasks, a sequence of consecutive utterances may jointly contain many important conversation-level phenomena such as topical coherence, lexical entrainment [23], and adjacency pairs [24] across utterances. It is therefore anticipated that if cross-utterance conversation-level information clues could be infused into language modeling, the performance of ASR *N*-best hypothesis reranking, will be considerably improved. Following this line of thought, we in this paper seek to integrate cross-utterances, long-term word interaction cues into the ASR *N*-best hypothesis reranking process, taking PBERT as the backbone LM, with two disparate fusion strategies: early fusion and late fusion.

*1) Early fusion:* For the early-fusion strategy, the BERT module is modified to take as input the concatenation of each hypothesis of the current utterance and the topmost ASR hypotheses (sequentially generated by *N*-best hypothesis reranking) of preceding m utterances with a special symbol [SEP] as the delimiter, as shown in Figure 3(a). We thus can generate a cross-utterance, history-aware embedding for each hypothesis of the current utterance to achieve enhanced oracle hypothesis prediction (denoted by EHPBERT). Note here that EHPBERT by nature performs word- or token-level fusion between the conversation history and each hypothesis of the current utterance.

*2) Late fusion:* Unlike the early-fusion strategy that modifies the BERT module to conduct cross-utterance, multi-head attention between words (or tokens), the late-fusion strategy attempts to factor in the semantic relationship between each hypothesis of the current utterance and those words that occur in the historical utterances so as to modulate the [CLS] embedding vector (viz. semantic representation) $\mathbf{h}_n$ of each hypothesis $n$, as illustrated in Figure 3(b). In doing so, a context embedding $\mathbf{c}_n^W$ is constructed by an attention mechanism ATT [25] that computes the similarity between the semantic representation of each hypothesis and the contextualized embeddings $E^W = (\mathbf{w}_1, ..., \mathbf{w}_L)$ of all words involved in the historical utterances:

$$\mathbf{c}_n^W = \text{ATT}(\mathbf{h}_n, E^W). \quad (1)$$

The resulting context embedding is in turn appended to the original semantic representation of each hypothesis of the current utterance to be fed in to the FNN module for oracle hypothesis prediction. We refer to this method as LHPBERT for short hereafter.

### C. Audio Fusion

It is anticipated to obtain enhanced ASR *N*-best hypothesis reranking by modeling cross-utterance, word-word interaction and semantic cues with either EHPBERT or LHPBERT. On a separate front, we believe that a reconsideration of the context relationship between the semantic representation of each hypothesis and the intermediate frame-wise acoustic embeddings $E^A = (a_1, ..., a_T)$ generated by the encoder module of an end-to-end (E2E) ASR system would provide extra discriminative information to additionally benefit *N*-best hypothesis reranking [26][27][28]. Figures 4 and 5 respectively show a schematic description of an audio-fusion strategy and its corresponding attention mechanism, while the attention mechanism can be succinctly expressed by

$$c_n^A = \text{ATT}'(h_n, E^A), \quad (2)$$
$$g_s = \text{Sigmoid}(W^c[c_n^A; h_n] + b_1), \quad (3)$$
$$h_n' = \text{Relu}(h_n + W^g[h_n \odot g_s] + b_2), \quad (4)$$

where $\odot$ represents the elementwise dot production and $W^c$, $W^g$, $b_2$ and $b_3$ are trainable parameters. More specifically, the functionality of Eqs. (2) and (3) is to model the context correlation between the sequence of acoustic representations $E^A$ and the semantic representation $h_n$ of each hypothesis $n$, producing a gating vector $g_s$ with the Sigmoid function. After that, Eq. (4) seeks to obtain a modulated semantic vector $h_n'$ with the gating vector $g_s$ and the Relu function in succession.

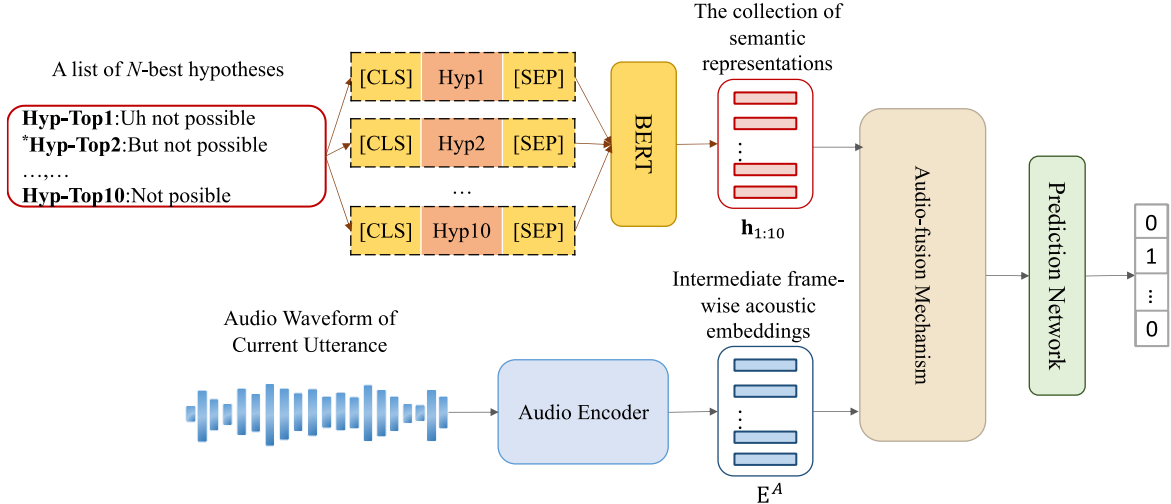

Fig.4. A schematic depiction of our audio-fusion strategy, which shows the entire pipeline for audio fusion with the PBERT model.

## III. EMPIRICAL EXPERIMENTS

### A. Experimental steup

We evaluate our proposed methods on the AMI meeting transcription benchmark corpus and task [29]. This speech corpus consisted of utterances collected with the individual headset microphones (IHM), while a pronunciation lexicon of 50K words was used. Table 1 shows some basic statistics of the AMI corpus for our experiments.

The baseline ASR system (which was employed to produce the *N*-best hypothesis list for a given utterance) adopted a Transformer-based ASR model[1] with the ESPnet toolkit [30]. The baseline LSTM-based LM and Transformer-based LM were also built with ESPnet. The LSTM-based LM consisted of 2 layers and 256 neurons for each[2]. Transformer-based LM consisted of 2 transformer blocks, where each block has 8 attention heads of dimensions 256 and 256 units in the feed-forward layers[3]. As to the various BERT-based LMs, we adopted the HuggingFace package [31] to develop them, using the bert-base-uncased model as the initial model. It had 12 Transformer blocks, each of which contained 12 multi-head self-attention operators and 768 hidden units. The BERT-based LMs were fine-tuned with 2 epochs and optimized with AdamW method [32]. In addition, the learning rate was set to 5e-5, the batch size was set to 16, and the gradient was accumulated 32 times for each update.

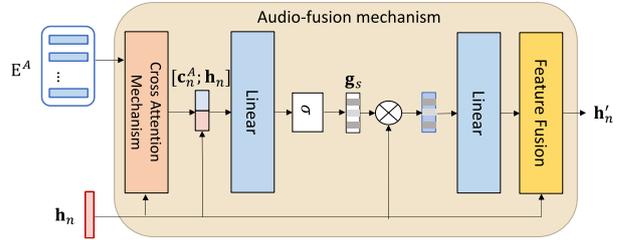

Fig.5. Illustration of our audio-fusion mechanism.

TABLE I. BASIC STATISTICS OF THE AMI CORPUS.

| Items | Train | Dev | Eval |
|---|---|---|---|
| # Hours | 78 | 8.71 | 8.97 |
| # Utters. | 108,104 | 13,059 | 12,612 |
| # Spk. Turns | 10,492 | 1,192 | 1,166 |
| # Words Per Utter. | 7.42 | 7.27 | 7.1 |

TABLE II. WORD ERROR RATE RESULTS ON THE EVALUATION SET FOR *N*-BEST HYPOTHESIS RERANKING WITH THE BASELINE LMS.

| Method | WER (%) | Latency (ms) |
|---|---|---|
| LSTM-based LM | 18.45 | 921.259 |
| Transformer-based LM | 18.58 | 878.847 |
| PBERT | **17.87** | **81.184** |

---

[1] https://zenodo.org/record/4615756#.YVaojW1BzFo
[2] espnet/egs2/librispeech/asr1/conf/tuning/train_lm_adam.yaml
[3] espnet/egs2/librispeech/asr1/conf/tuning/train_lm_transformer2.yaml

TABLE III. WORD ERROR RATE RESULTS ACHIEVED BY THE EARLY-FUSION AND LATE-FUSION STRATEGIES, AS WELL AS THEIR COMBINATION.

| Method | WER (%) | Latency (ms) |
|---|---|---|
| **Early fusion** | | |
| EHPBERT (ν = 1) | 17.75 | 120.096 |
| EHPBERT (ν = 2) | **17.72** | **151.936** |
| EHPBERT (ν = 3) | 17.73 | 197.504 |
| **Late fusion** | | |
| LHPBERT (ω = 5) | 17.77 | 93.792 |
| LHPBERT (ω = 10) | **17.73** | **95.328** |
| LHPBERT (ω = 15) | 17.75 | 97.648 |
| **Early fusion + Late fusion** | | |
| EHPBERT + LHPBERT (ν = 2, ω = 10) | 17.64 | 154.272 |

TABLE IV. WORD ERROR RATE RESULTS ACHIEVED BY AUDIO FUSION IN CONJUNCTION VARIOUS BERT-BASED LMS.

| Method | WER (%) | Latency (ms) |
|---|---|---|
| **Audio fusion** | | |
| PBERT | 17.76 | 83.024 |
| EHPBERT (ν = 2) | 17.69 | 152.992 |
| LHPBERT (ω = 10) | 17.72 | 97.024 |
| EHPBERT + LHPBERT (ν = 2, ω = 10) | **17.56** | **166.544** |

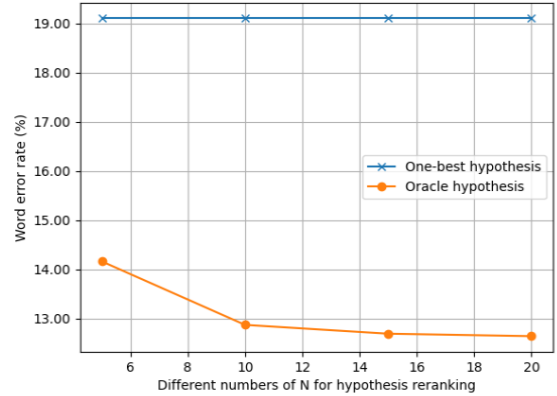

Fig. 6. A depiction of the WER results of the ASR system (top-one) and those of the oracle hypothesis as a function of the size of the N-best list.

The reranking performance and latency on the test set were evaluated based on the average word error rate (WER) and response time for the test utterances, respectively. The rescoring process was performed on a single Nvidia TITAN RTX GPU and the latency indicates that the time required to accomplish the reranking process after the ASR model generates the *N*-best hypothesis list for a given test utterance.

### B. Experimental result

At the outset, we report on the word error rate (WER) result obtained by the ASR system and those of the oracle hypothesis in the *N*-best list. As the graphical depiction shown in Figure 6, the top-one hypothesis of the ASR system is 19.10%. In addition, the WER result of the oracle hypothesis is 12.87% when *N* is set equal to 10, whereas increasing the value of *N* only leads to a slight WER reduction of the oracle hypotheses (when *N* > 10). As such, we will set *N*=10 for the following ASR *N*-best hypothesis reranking experiments, unless stated otherwise.

We then assess the efficacy of rescoring methods, the baseline *N*-best hypothesis results achieved by the LSTM-based LM, the Transformer-based LM, and PBERT are shown in Table 2. As can be seen from Table 2, all of these LM methods can provide substantial WER reduction when compared to the top-one result of the ASR system (with WER of 19.10%). PBERT outperforms the LSTM-based LM by a margin of 3.14% relative WER reduction, and provides a relative WER reduction of 3.82% over the Transformer-based LM. In terms of the latency for ASR *N*-best hypothesis reranking, PBERT is dramatically faster than the other two LM methods due to its one-shot execution process. More specifically, PBERT is about 10x and 11x faster than the Transformer-based LM and the LSTM-based LM, respectively.

In the second set of experiments, we turn to the evaluation of the effectiveness of cross-utterance language modeling with the two fusion strategies proposed in this paper, viz. early function and late function (cf. Section II-B). Their corresponding WER results are shown in Table III. Inspection of Table III reveals several noteworthy points. First, early fusion (denoted by EHPBERT) and late fusion (denoted by LHPBERT) perform in line with each other, while LHPBERT is much faster than EHPBERT. Second, early fusion has its best WER result when two immediately preceding utterances (ν=2) are taken into consideration, while late fusion delivers its best WER result when 10 immediately preceding words in the conversation history (ω=10) are used. Third, combination of early fusion and late fusion can yield an additional improvement, leading to an WER result of 17.64%.

In the last set of experiments, we assess the utility of integrating audio fusion (cf. Section II-C) into the various BERT-based LM proposed in this paper. The corresponding results are shown in Table IV. A closer look at Table IV reveals at least two things. First, the additional adoption of audio fusion can provide extra gains for all variants of our BERT-based LMs, although the corresponding execution latencies increase as compared to that of their counterparts without audio fusion. Second, when combining these three fusion strategies (viz. late fusion, late fusion and audio fusion) all together, we can obtain the best WER result of 17.56%, which ultimately offers relative WER improvements of 4.82% and 1.73% in relation to LSTM and PBERT, respectively.

### IV. CONCLUSION AND FUTURE WORK

In this paper, we have explored different fusion strategies, viz. early fusion, late fusion and audio fusion, for ASR *N*-best hypothesis reranking. These fusion strategies work in tandem to infuse cross-utterance word interaction and semantic cues, as well as the relatedness between the semantic embedding of each hypothesis and the acoustic embeddings the corresponding test utterance, into language modeling. A series of empirical evaluations on the AMI meeting transcription benchmark task seem to confirm the practical utility of our methods in comparison to de facto standard autoregressive LM methods. In future work, we plan to investigate more effective attention mechanisms to infuse various lexical, semantic and acoustic cues into language modeling for ASR. It is also worthwhile to further validate our LM methods on more diverse ASR tasks, such as TED-LIUM [33], LibriSpeech [34] and others.

## REFERENCES

[1] M. Ravanelli et al., "SpeechBrain: A general-purpose speech toolkit," arXiv preprint arXiv:2106.04624, 2021.

[2] L. Dong et al., "Speech-transformer: a no-recurrence sequence-to-sequence model for speech recognition," in Proceedings of IEEE International Conference on Acoustics, Speech and Signal Processing (ICASSP), pp. 5884–5888, 2018.

[3] A. Shenoy et al., "Adapting long Ccontext NLM for ASR rescoring in conversational agents," in Proceedings of the Annual Conference of the International Speech Communication Association (INTERSPEECH), pp. 3246–3250, 2021.

[4] W. Xiong et al., "Session-level language modeling for conversational speech," in Proceedings of the conference on Empirical Methods in Natural Language Processing (EMNLP), pp. 2764–2768, 2018.

[5] S. Yang, N. Yu, G. Fu, "A discourse-aware graph neural network for emotion recognition in multi-party conversation," in Proceedings of the conference on Empirical Methods in Natural Language Processing (EMNLP), pp. 2949–2958, 2021.

[6] E. Grave et al., "Improving neural language models with a continuous cache," in Proceedings of International Conference on Learning Representations (ICLR), 2017.

[7] B. Krause et al., "Dynamic evaluation of transformer language models," arXiv preprint arXiv:1904.08378v1, 2019.

[8] G. Sun et al., "Transformer language models with LSTM-based cross-utterance information representation," in Proceedings of IEEE International Conference on Acoustics, Speech and Signal Processing (ICASSP), pp. 7363–7367, 2021.

[9] Z. Dai et al., "Transformer-XL: Attentive language models beyond a fixed-length context," in Proceedings of the Annual Meeting of the Association for Computational Linguistics (ACL), pp. 2978–2988, 2019.

[10] X. Geng, X. Feng, B. Qin, "Learning to rewrite for non-autoregressive neural machine translation," in Proceedings of the conference on Empirical Methods in Natural Language Processing (EMNLP), pp. 3297–3308, 2021.

[11] A. Mitrofanov, et al. "LT-LM: a novel non-autoregressive language model for single-shot lattice rescoring," in Proceedings of the Annual Conference of the International Speech Communication Association (INTERSPEECH), pp. 4039–4042, 2021.

[12] U. Khandelwal, H. He, P. Qi, D. Jurafsky, "Sharp nearby, fuzzy far away: How neural language models use context," in Proceedings of the Annual Meeting of the Association for Computational Linguistics (ACL), 2018.

[13] J. Devlin et al., "BERT: Pre-training of deep bidirectional Transformers for language understanding," in Proceedings of the Conference of the North American Chapter of the Association for Computational Linguistics: Human Language Technologies (NAACL-HLT), pp. 4171–4186, 2019.

[14] T. Mikolov et al., "Recurrent neural network based language model," in Proceedings of the Annual Conference of the International Speech Communication Association (INTERSPEECH), pp. 1045-1048, 2010.

[15] M. Sundermeyer et al., "LSTM neural networks for language modeling," in Proceedings of the Annual Conference of the International Speech Communication Association (INTERSPEECH), 2012.

[16] E. Beck, R. Schluter, and H. Ney, "LVCSR with transformer language models," in Proceedings of the Annual Conference of the International Speech Communication Association (INTERSPEECH), pp. 1798–1802, 2020.

[17] S.-H. Chiu and B. Chen, "Innovative BERT-based reranking language models for speech recognition," in Proceedings of the IEEE Spoken Language Technology Workshop (SLT), pp. 266–271, 2021.

[18] S.-H. Chiu et al., "Cross-sentence neural language models for conversational speech recognition," in Proceedings of the IEEE International Joint Conference on Neural Networks (IJCNN), pp. 1–7, 2021.

[19] S.-H. Chiu, T.-H. Lo, F.-A. Chao, B. Chen, "Cross-utterance reranking models with BERT and graph convolutional networks for conversational speech recognition," arXiv preprint arXiv:2106.06922, 2021.

[20] J. Salazar et al., "Masked language model scoring," in Proceedings of the Annual Meeting of the Association for Computational Linguistics (ACL), pp. 2699–2712, 2020.

[21] J. Shin, Y. Lee, and K. Jung, "Effective sentence scoring method using BERT for speech recognition," in Proceedings of The Asian Conference on Machine Learning (ACML), pp. 1081–1093, 2019.

[22] A. Jain et al., "Finnish ASR with deep transformer models," in Proceedings of the Annual Conference of the International Speech Communication Association (INTERSPEECH), pp. 3630–3634, 2020.

[23] S. E. Brennan and H. H. Clark, "Conceptual pacts and lexical choice in conversation," Journal of Experimental Psychology: Learning, Memory, and Cognition, pp.1482–1493, 1996.

[24] E. A. Schegloff, "Sequencing in conversational openings," American Anthropologist, pp.1075–1095, 1968.

[25] J. K. Chorowski et al., "Attention-based models for speech recognition," in Advances in neural information processing systems (NeurIPS), pp. 577–585, 2015.

[26] G. Zheng et al. "Wav-bert: Cooperative acoustic and linguistic representation learning for low-resource speech recognition," in Proceedings of the Annual Meeting of the Association for Computational Linguistics (ACL), pp. 2766–2777, 2021.

[27] S. Zhang et al. "End-to-end spelling correction conditioned on acoustic feature for code-switching speech recognition," in Proceedings of the Annual Conference of the International Speech Communication Association (INTERSPEECH), pp. 266-270, 2021.

[28] Y. Zhu et al. "WavBERT: Exploiting semantic and non-semantic speech using wav2vec and bert for dementia detection," in Proceedings of the Annual Conference of the International Speech Communication Association (INTERSPEECH), pp. 3790-3794, 2021.

[29] J. Carletta et al., "The AMI meeting corpus: A pre-announcement," in Proceedings of the International Workshop on Machine Learning for Multimodal Interaction, pp. 28–39, 2005.

[30] S. Watanabe, et al., "ESPnet: End-to-end speech processing toolkit," in Proceedings of the Annual Conference of the International Speech Communication Association (INTERSPEECH), pp. 2207–2211, 2019.

[31] T. Wolf, et al, "Transformers: State-of-the-art natural language processing," in Proceedings of the Conference on Empirical Methods in Natural Language Processing: System Demonstrations (EMNLP), pp. 38–45, 2020.

[32] I. Loshchilov and F. Hutter, "Decoupled weight decay regularization," in International Conference on Learning Representations (ICLR), 2019.

[33] F. Hernandez et al. "TED-LIUM 3: twice as much data and corpus repartition for experiments on speaker adaptation." In Proceedings of the International conference on speech and computer (SPECOM), pp. 198–208, 2018.

[34] V. Panayotov et al. "Librispeech: an asr corpus based on public domain audio books," in Proceedings of the Annual Conference of the International Speech Communication Association (INTERSPEECH), pp. 5206–5210, 2015.